# GDPR-Compliant Person Recognition in Industrial Environments Using MEMS-LiDAR and Hybrid Data

Dennis Basile[a,*], Dennis Sprute[a], Helene Dörksen[b], Holger Flatt[a]

*[a]Fraunhofer IOSB, Industrial Automation Branch (IOSB-INA), 32657 Lemgo, Germany*
*[b]OWL University of Applied Sciences and Arts, 32657 Lemgo, Germany*

* Corresponding author. Tel.: +49 5261 9429022. *E-mail address:* dennis.basile@iosb-ina.fraunhofer.de

## Abstract

The reliable detection of unauthorized individuals in safety-critical industrial indoor spaces is crucial to avoid plant shutdowns, property damage, and personal hazards. Conventional vision-based methods that use deep-learning approaches for person recognition provide image information but are sensitive to lighting and visibility conditions and often violate privacy regulations, such as the General Data Protection Regulation (GDPR) in the European Union. Typically, detection systems based on deep learning require annotated data for training. Collecting and annotating such data, however, is highly time-consuming and due to manual treatments not necessarily error free. Therefore, this paper presents a privacy-compliant approach based on Micro-Electro-Mechanical Systems LiDAR (MEMS-LiDAR), which exclusively captures anonymized 3D point clouds and avoids personal identification features. To compensate for the large amount of time required to record real LiDAR data and for post-processing and annotation, real recordings are augmented with synthetically generated scenes from the CARLA simulation framework. The results demonstrate that the hybrid data improves the average precision by 44 percentage points compared to a model trained exclusively with real data while reducing the manual annotation effort by 50 %. Thus, the proposed approach provides a scalable, cost-efficient alternative to purely real-data-based methods and systematically shows how synthetic LiDAR data can combine high performance in person detection with GDPR compliance in an industrial environment.

*Keywords*: Person detection; 3D point cloud; MEMS-LiDAR; Synthetic data; Real data

## 1. Introduction

In industrial environments, both indoors and outdoors, the recognition of individuals is crucial for safety monitoring and analysis. Current research focuses on developing reliable and GDPR-compliant systems [1]. These systems must operate in real time and comply with privacy regulations, while functioning effectively under varying weather and lighting conditions. MEMS-LiDAR sensors present a promising solution for GDPR-compliant person recognition by generating precise 3D point clouds [2]. Unlike rotating LiDAR systems, which are commonly used in autonomous driving and typically offer 360° coverage, MEMS-LiDAR sensors are more compact, mechanically robust, costs are lower and offer a higher spatial resolution and faster sampling rates within a limited field of view [3]. These characteristics make them particularly well-suited for static industrial applications, where detailed monitoring of specific zones is required and full-environment scanning is not necessary. The higher point density within the sensor's field of view allows better detection of small body parts, such as hands, which is especially beneficial for safety monitoring [2]. Additionally, since LiDAR sensors do not capture personal features, they reduce the risk of individual identification and ensure privacy. However, there is a lack of datasets for MEMS-LiDAR, which complicates the development of robust neural networks [2]. Obtaining a sufficient number of real-world data is both costly and time-consuming, as it involves expensive equipment and extensive measurement across diverse conditions. This includes setup tasks such as installation, alignment, and post-processing of data.

Although datasets for rotating LiDAR sensors exist, they are not directly applicable to MEMS-LiDAR systems due to differences in sensor characteristics, field of view, and point cloud distribution [4]. Therefore, new datasets have to be created to meet the specific requirements of MEMS-LiDAR applications. However, a promising and cost-effective alternative for dataset creation exists through the use of simulated data generated with platforms like CARLA, which offer a flexible solution by replicating complex scenarios and varying environmental conditions [5].

This paper aims to enhance recognition performance in industrial indoor and outdoor settings by a new approach of fusing real and simulated data. To achieve this, two specific sub-goals are addressed:

1. **Creation of a new MEMS-LiDAR dataset for neuronal network training:**
   Development of a combined dataset consisting of real and synthetic LiDAR data, collected in public environments for neural network training.

2. **Evaluating the impact of combination of real and synthetic LiDAR data:**
   Analysis of how the combination of real and synthetic LiDAR data affect classification and localization performance. This includes a systematic evaluation of different training data configurations to assess how well the models generalize to unseen industrial scenarios.

This paper is organized as follows: Section 2 reviews related work. In section 3, the approach for creating real and synthetic datasets is described. Section 4 describes the evaluation, including the network architecture, the used hyperparameters and the training process. In section 5, the results of the case study are presented. Section 6 discusses the results and section 7 concludes the paper.

## 2. Related Work

Deep-learning-based person recognition in 3D-LiDAR point clouds relies heavily on extensive and representative datasets. Publicly available benchmark datasets mainly focus on mechanically rotating LiDAR sensors, especially the widely used Velodyne HDL-64E [2]. These datasets primarily serve autonomous driving applications and differ significantly from industrial indoor and outdoor scenarios for person detection. Moreover, the technical characteristics of rotating LiDAR systems such as their 360° field of view and distinct scanning patterns limit their direct applicability to newer MEMS-based LiDAR systems, which typically feature higher resolution, a smaller field of view, and different scanning behaviors [6].

Currently, optimized datasets for MEMS-based LiDAR are lacking, posing challenges for developing robust neural networks tailored to these sensor types [2]. Real-world data collection using MEMS-based LiDAR sensors is expensive, labor-intensive, and requires extensive preprocessing and annotation. To address this shortage, recent research has increasingly explored synthetic data generation methods.

Generative models represent one approach, typically creating artificial point clouds from images or text descriptions [7, 8]. Despite advances, these generative methods mainly generate isolated objects rather than complex scenes, which limits their suitability for training robust person recognition networks that require diverse contextual scenarios.

As an alternative, physics-based simulations using software platforms like CARLA [5], NVIDIA DRIVE Sim [9] or Matlab/Simulink [10] have gained popularity for realistically simulating complete sensor environments. CARLA, specifically, stands out as an open-source, flexible platform that enables custom sensor modeling through its Python API [5]. Initial studies demonstrated the feasibility of MEMS-based LiDAR simulations in CARLA, but focused exclusively on geometric accuracy [11]. However, no systematic method has yet been proposed to evaluate the effectiveness of simulated MEMS-based LiDAR data in supporting real-world data for person recognition tasks.

This paper directly addresses this gap by systematically combining simulated and real MEMS-LiDAR data, evaluating their collective impact on neural network training and person detection performance in realistic industrial contexts.

## 3. Creation of Real and Synthetic Datasets for Person Detection

An overview of the dataset creation process for both synthetic and real data is illustrated in Fig. 1. The dataset created for this work consists of both real and synthetic point clouds that were generated using two different approaches. All datasets are annotated with a single object class called "person". The real data was recorded using a 3D-LiDAR and required time-consuming manual annotation to ensure accurate object labels. In comparison, the synthetic dataset was created using a simulation environment that enabled automatic annotation during data generation.

### 3.1. Real 3D Point Cloud Dataset

The real 3D-LiDAR point cloud dataset was created using a MEMS-based LiDAR sensor, which is the Cube1 sensor from the manufacturer Blickfeld [13]. This sensor can be parameterized according to application requirements. To ensure reliable recognition of people, a resolution of 200 scan lines and a field of view of 72° horizontally and 30° vertically was selected [2].

The data collection took place under sunny skies and light rain on the grounds of the Fraunhofer IOSB-INA Institute [18]. The LiDAR sensor was mounted statically at a height of 4 meters with a tilt of 16° to capture the entire field of view of the area. Various persons were recorded.

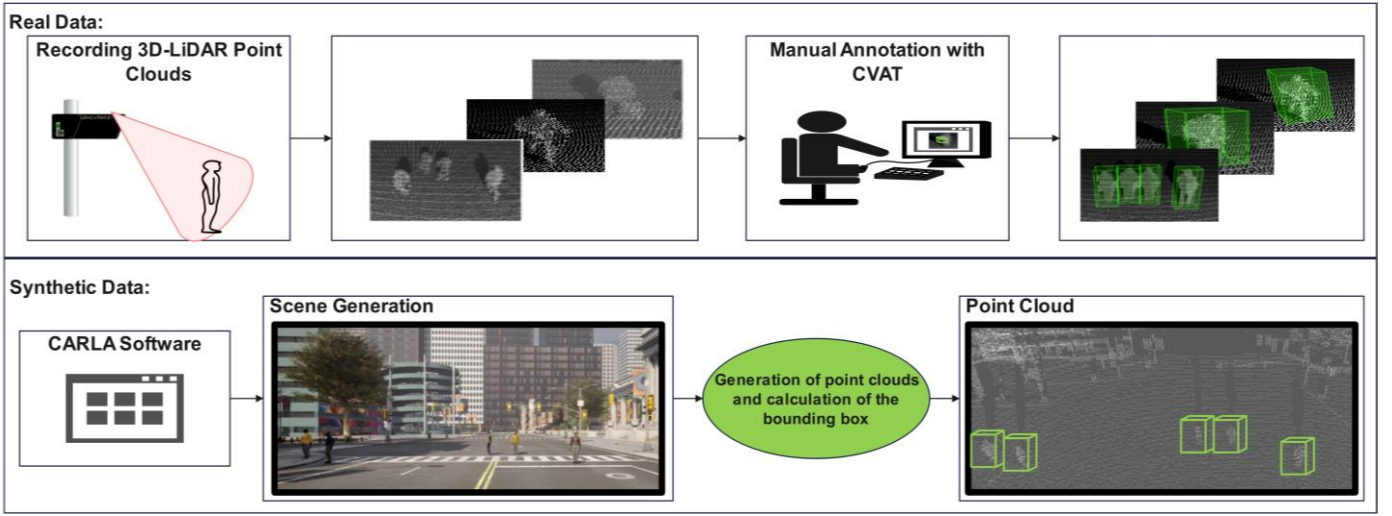

Fig. 1. Overview of the proposed approach for creating synthetic and real datasets

After capturing the different persons, the data was prepared for training a neural network for direct 3D point cloud processing. The collected data was recalculated to the origin, and the tilt and height were removed. After normalization, the data was manually annotated using the CVAT tool [14] for marking 3D bounding boxes. The bounding boxes contain the coordinates of the center of the bounding box, specified as $x$, $y$, $z$, along with the width $dx$, the length $dy$, and the height $dz$. The dataset focuses on the recognition of people, so accompanying objects were not annotated, only the individuals.

The real dataset includes 2,100 annotated point clouds specifically created for person recognition with MEMS-LiDAR.

### 3.2. 3D Point Cloud Generation Using CARLA Simulation

To generate the simulation dataset, the CARLA software version 9.15, based on Unreal Engine 4 was used [5]. CARLA provides realistic visualization and simulation capabilities for diverse traffic scenarios. By employing CARLA's Python API, a MEMS-based LiDAR can be integrated and simulated within this environment. Initial approaches for projecting a MEMS-LiDAR were presented by Berens et al. [11]. Building upon their approach, this paper demonstrates how annotated datasets for person detection can be automatically generated. Calculating a LiDAR measurement point from the simulated scene involves three primary steps:

**1. Scanning Procedure:**
Horizontal and vertical mirror movements of the MEMS-LiDAR sensor are simulated using sine and cosine functions at a defined scan frequency, calculating the (x, y) positions of points within the virtual scene.

**2. Depth Information Extraction:**
Depth information is obtained directly from an integrated virtual depth camera within the CARLA environment.

### 3. Transformation into 3D Coordinates:
Using the image resolution from the depth camera and field-of-view parameters, the inverse intrinsic matrix is computed. This matrix allows the conversion of 2D scan points (x, y) into 3D coordinates. By combining these coordinates with the extracted depth data, the final 3D coordinates (x, y, z) for each measurement point are determined. Detailed formulas and methods for calculating coordinates and generating synthetic point clouds are thoroughly described in the work of Berens et al. [11].

For automatic person annotation, virtual persons were dynamically spawned using the CARLA API, moving along predefined trajectories within the sensor's fields of view. Since persons in CARLA are implemented as instances of a person class, their precise positions can be directly retrieved using the API. This enables automatic annotation through 3D bounding boxes. The transformation into sensor coordinates is calculated by:

$$P_{sensor} = T_{sensor}^{-1} \cdot P_{world}$$

$P_{world}$ represents the global corner points of each bounding box, which are converted into sensor coordinates via the inverse sensor transformation $T_{sensor}^{-1}$. This process makes it easier to clearly assign persons to their corresponding bounding boxes in each generated point cloud. Using this method, 2,100 annotated point clouds were generated at three person crossings and one person pathway location, covering various scenarios with between 3 and 10 persons each. This range was chosen because larger groups result in excessive occlusions, and it roughly corresponds to the maximum number of individuals that the sensor's field of view can reliably capture. To ensure a fair performance comparison between synthetic and real datasets, the training set size was consistently fixed at 2,100 samples for each experiment.

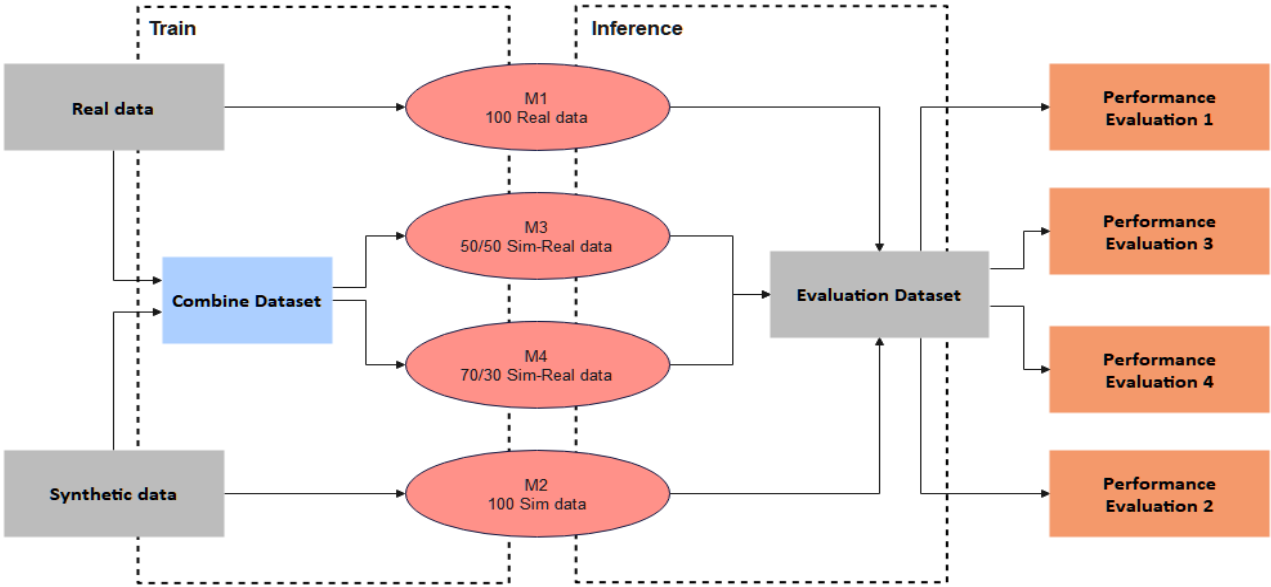

Fig. 2 Overview of the evaluation pipeline

The data show high variability in human movement and body orientation. In addition, occlusions occur frequently and the spatial distribution of persons varies considerably, i.e. people move side-by-side, move in groups, or cross paths. This variety of scenarios, as commonly observed in urban environments, makes the dataset valuable not only for urban applications, but also for validating people detection and tracking algorithms in industrial environments.

### 3.3. Novel Real Evaluation Dataset

An additional test dataset was created to evaluate the detection performance on unseen scenarios. This extension was necessary because both the existing real dataset and the simulated dataset exclusively consist of scenes from public areas. Consequently, neither dataset contains industrial scenarios. The aim was to investigate how the trained deep-learning model, previously trained on these two datasets, generalizes to unfamiliar scenarios and to assess its recognition robustness in such conditions.

For this purpose, a MEMS-LiDAR sensor was mounted in a real industrial environment at the Smart Factory OWL in Lemgo at a height of 5 meters, tilted at an angle of 23° to fully capture the entrance area. The test dataset comprises approximately 230 manually annotated point clouds.

## 4. Evaluation

This section addresses the second objective of this work: Neural networks are trained using different data compositions to analyze how well the models generalize to unseen industrial environments. Fig. 2 illustrates the overall evaluation pipeline, from dataset composition to model training and final testing on industrial scenarios. Two hybrid datasets are investigated: A sim-real ratio of 50/50 is chosen to provide a balanced dataset as a training base, while a sim-real ratio of 70/30 tests the model's behavior with a higher proportion of simulation data to evaluate its robustness and adaptability. Furthermore, this section describes the experimental setup, including the training configurations, the structure of the evaluation pipeline, and the hardware used for training and testing.

### 4.1. SECOND Deep Neural Network

The detection and evaluation of persons in point clouds were performed using the SECOND (Sparsely Embedded Convolutional Detection) deep-learning network. SECOND is a voxel-based method that partitions point clouds into predefined voxels and subsequently forwards these voxels to a neural network for automatic feature extraction [12].

The choice of this architecture was motivated by the comparative analysis conducted by Basile et al., who evaluated various state-of-the-art deep-learning approaches for direct point cloud processing [2]. Their findings indicated that SECOND provides an effective balance between inference speed and recognition accuracy.

### 4.2. Training

For training and executing the deep learning architecture SECOND, the Open-Source Point Cloud Detection (OpenPCDet) framework was utilized [15]. The deep learning architectures were trained on a Windows system with the following specifications: AMD Ryzen 9 3900X 12-core processor with 64 GB DDR4 RAM and an RTX 2080 graphics card with 8 GB of memory.

To ensure reliable results, each of the four datasets, comprising the real dataset, the sim dataset, and the two sim-real datasets, was randomly divided into 70% training data and 30% validation data. Various data augmentation techniques, such as rotation, scaling, and mirroring of the point cloud, were employed to artificially expand the dataset and enhance the performance of the trained model. Additionally, the following hyperparameters were selected based on recommendations from [2]: the optimization algorithm Adam, combined with the OneCycleLR learning rate scheduler [16], an initial learning rate of $10^{-4}$, and a total of 120 training epochs.

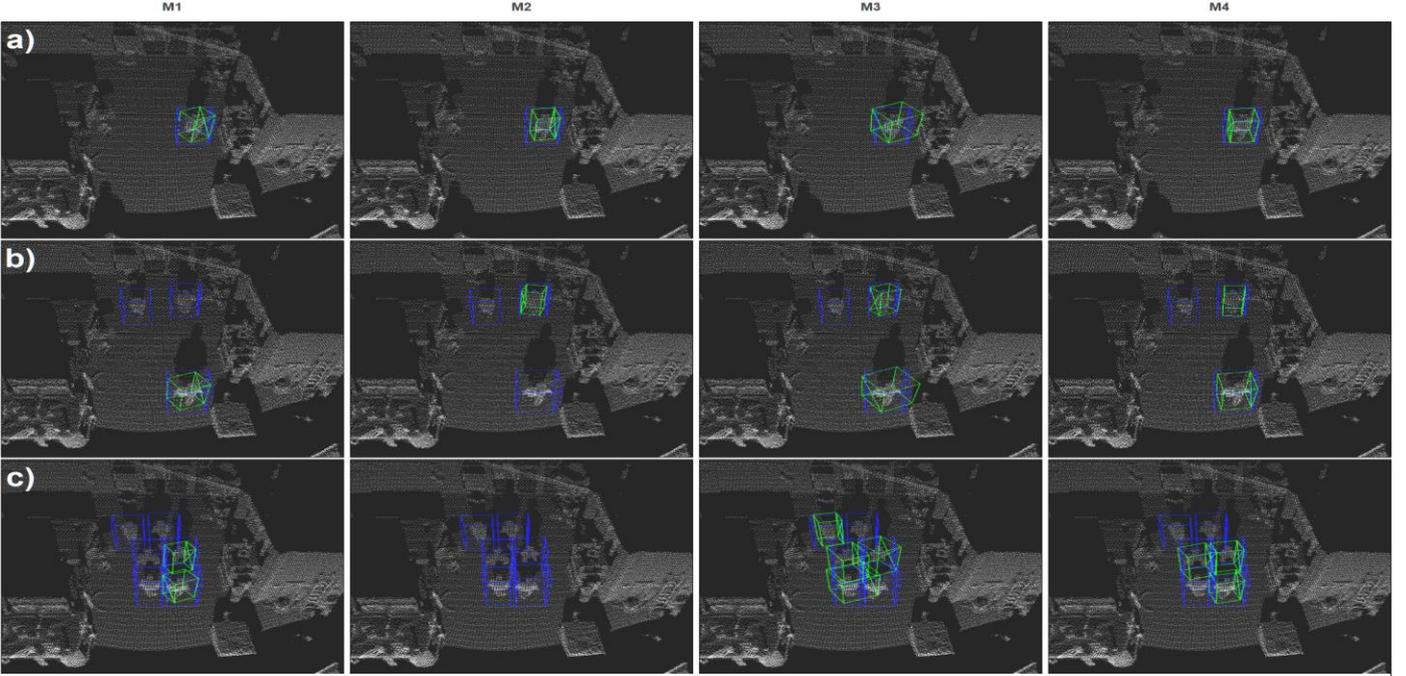

Fig. 3: Examples of classification results of models M1 to M4. The green bounding boxes represent the output of the deep learning network, while the blue bounding boxes correspond to the ground truth bounding boxes derived from the annotation.

### 4.3. Test setup and design

Four neural networks based on the SECOND architecture were trained with identical hyperparameters but different compositions of synthetic and real training data (Tab. 2).:

Table 2. Neural network model configurations based on SECOND architecture

| Model ID | Synthetic / Real (%) | Synthetic / Real (Count) |
|----------|----------------------|--------------------------|
| M1 | 0 / 100 | 0 / 2100 |
| M2 | 100 / 0 | 2100 / 0 |
| M3 | 50 / 50 | 1050 / 1050 |
| M4 | 70 / 30 | 1470 / 630 |

The performance of the trained neural networks is subsequently evaluated on the dataset described in Sec. 3.3, which was collected specifically for industrial scenarios. The Average Precision (AP) [17] is used for evaluation with an Intersection over Union (IoU) threshold of 0.5. Following evaluation, each model yields an AP score that quantifies its recognition performance and provides the basis for comparison with the other models.

## 5. Results

The results of the evaluation of the four models M1-M4 on the evaluation dataset with industry scenes are summarized in Table 3. Some qualitative classification and localization results of the individual models are visualized in Figure 3.

Table 3. SECOND performance: training on different datasets (M1, M2, M3, M4) and prediction on an evaluation dataset for unseen industry scenes

| Model ID | Architecture | AP (IoU = 0.5) |
|----------|--------------|----------------|
| M1 | SECOND | 0.10 |
| M2 | SECOND | 0.27 (+17 %) |
| M3 | SECOND | 0.54 (+44 %) |
| M4 | SECOND | 0.45(+35 %) |

Model M1 achieves a recognition performance of only 10 % in the evaluation dataset, while model M3 reaches a value of 54 %. In comparison with model M1, this indicates a relative increase of 44 % for model M3 and 35 % for model M4 in terms of recognition performance. Using only simulated data (model M2) improves the recognition performance by 17 %. Fig. 3 illustrates the results of the four models: In row a), all models correctly classify the persons shown. Fig. 3 row b) shows that models M1 and M2 show significant deviations from the ground truth, while models M3 and M4 better match their predicted bounding boxes to the ground truth. If there are several people in the nearby area, model M1 recognizes two out of six objects, model M2 no object, model M3 four out of six objects and model M4 three out of six objects (row c)).

## 6. Discussion

The results demonstrate that the research question "*How does the combination of real and synthetic LiDAR data affect classification and localization performance?*" can be successfully responded. Based on the achieved AP, it is clear that models M3 and M4, through the mixture of real and synthetic data, ensure robust object detection in unknown scenarios. Model 3 achieves an improvement of 44 % compared to the model trained on the real dataset, which only reaches an AP of 10 %. Such a result is poor when considering the significant effort invested in recording point clouds and manual annotation. The poor result on the real-world dataset may be due to the lack of complex scenes, as the data was captured from a single viewpoint and includes less situations with many people or close interactions. In contrast, the simulated dataset specifically includes such complex scenes, which helped the model generalize better. The gap in scene diversity likely had a significant impact on performance. Nevertheless, a strong increase in recognition performance is observed, attributable to the combined training dataset. By simulating rare or difficult scenarios, as well as different angles and occlusions of objects, the robustness and generalization capability with respect to real recordings were enhanced. Furthermore, scenes like in Fig. 3, row c) show that when multiple people are present, and the shadows of the front persons partially occlude the point cloud, the M3 model performs best compared to M1 with only real data or M2 with only simulation data. Through this performance improvement with simulation data, it becomes apparent that even when simple real data are collected without difficult scenes, such as occluded persons, the classification performance can still be increased using simulation data that include these cases. Additionally, using model M3 as an example with a ratio of 50 % real data and 50 % synthetic data, only half the time is needed for collecting real data and annotation, which can significantly reduce effort and costs. For the intended industrial application, the current recognition rate of 54 % is still insufficient, as the model has so far not been trained with data from real-world scenes that reflect the actual application perspective. Including such data, which is representative of the application context, offers considerable potential for improving recognition performance. However, real application data was deliberately not used to first analyze the isolated influence of synthetic training data on recognition performance. Additionally, multi-frame tracking methods could further enhance recognition stability by leveraging temporal information across consecutive frames.

## 7. Conclusions and Future Work

In summary, it was shown that combining real and synthetic LiDAR data significantly improves the performance of neural networks for person detection in industrial environments. Specifically, mixing real and synthetic data at an equal rate leads to a substantial increase of 44 % in AP compared to models trained only with real data. Using simulation data allows for covering rare or complex scenarios, thus increasing the robustness and generalization ability of the models. This enables more efficient and cost-effective development of detection systems, as the effort for capturing and annotating real data is reduced. Future work could focus on an in-depth analysis of the distribution ratio between real and synthetic data in the dataset to further enhance performance. Another approach could be to train a model on a dataset with simulation data and perform fine-tuning with real data scenarios to boost performance. Additionally, domain adaptation methods could be used to further improve performance and enhance applicability in various industrial scenarios.

## References

[1] Sprute D, Westerhold T, Hufen F, Flatt H, Gellert F. DSGVO-konforme Personendetektion in 3D-LiDAR-Daten mittels Deep Learning Verfahren. In: Bildverarbeitung in der Automation (in German). Berlin: Springer; 2023. p. 33-45.

[2] Basile D, Sprute D, Dörksen H, Flatt H. Evaluation of 3D-LiDAR based person detection algorithms for edge computing. In: Image Processing Forum 2024. Karlsruhe: KIT Scientific Publishing; 2024. p. 159-170.

[3] Fernandes D, Silva A, Névoa R, Simões C, Gonzalez D, Guevara M, Novais P, Monteiro J, Melo-Pinto P. Point-cloud based 3D object detection and classification methods for self-driving applications: A survey and taxonomy. Information Fusion 2021;68: p. 161-191.

[4] Geiger A, Lenz P, Stiller C, Urtasun R. Vision meets robotics: The KITTI dataset. The International Journal of Robotics Research 2013;32: p. 1231-1237.

[5] Dosovitskiy A, Ros G, Codevilla F, Lopez A, Koltun V. CARLA: An open urban driving simulator. In: Proceedings of the 1st Annual Conference on Robot Learning. 2017;78: p. 1-16.

[6] Li Y, Shi H. Advanced Driver Assistance Systems and Autonomous Vehicles: From Fundamentals to Automation. Singapore: Springer; 2022.

[7] Fan H, Su H, Guibas L. A point set generation network for 3D object reconstruction from a single image. In: IEEE Conference on Computer Vision and Pattern Recognition (CVPR); 2017. p. 2463-2471.

[8] Nichol A, Jun H, Dhariwal P, Mishkin P, Chen M. Point-E: A system for generating 3D point clouds from complex prompts. arXiv preprint arXiv:2212.08751; 2022.

[9] NVIDIA Developer. NVIDIA DRIVE Sim [Internet]. Available from: https://developer.nvidia.com/drive/simulation. Accessed: Apr. 8, 2025.

[10] MathWorks. Automated Driving Toolbox [Internet]. Available from: https://de.mathworks.com/products/automated-driving.html. Accessed: Apr. 8, 2025.

[11] Berens F, Reischl M, Elser S. Generation of synthetic point clouds for MEMS LiDAR sensor. TechRxiv; 2022 Apr 27.

[12] Yan Y, Mao Y, Li B. SECOND: Sparsely embedded convolutional detection. Sensors 2018;18(10): p. 3337.

[13] Blickfeld Company. Cube 1 Blickfeld [Internet]. Available from: http://docs.blickfeld.com/cube/latest/index.html. Accessed: Apr. 15, 2025.

[14] CVAT.ai Corporation. Computer Vision Annotation Tool (CVAT) [Internet]. Available from: https://github.com/cvat-ai/cvat. Accessed: Apr. 15, 2025.

[15] OpenPCDet Development Team. OpenPCDet: An open-source toolbox for 3D object detection from point clouds [Internet]. GitHub repository; 2020. Available from: https://github.com/open-mmlab/OpenPCDet. Accessed: Apr. 25, 2025.

[16] Smith LN, Topin N. Super-convergence: very fast training of neural networks using large learning rates. In: Artificial Intelligence and Machine Learning for Multi-Domain Operations Applications. Proc SPIE 11006; 2019.

[17] Simonelli A, Bulò SR, Porzi L, Antequera ML, Kontschieder P. Disentangling monocular 3D object detection: From single to multi-class recognition. IEEE Trans Pattern Anal Mach Intell 2022;44(3): p. 1219-1231.

[18] Sprute D, Hufen F, Westerhold T, Flatt H. 3D-LiDAR-based pedestrian detection for demand-oriented traffic light control. In: IEEE 21st International Conference on Industrial Informatics (INDIN); 2023. p. 1-7.